\documentclass[conference]{IEEEtran}

\usepackage{graphicx}
\ifCLASSINFOpdf
\else
\fi

\usepackage{booktabs} 

\begin{document}
%
\title{Neuromorphic Visual Odometry System for Intelligent Vehicle Application with Bio-inspired Vision Sensor}

\author{\IEEEauthorblockN{Dekai Zhu, Jinhu Dong, Zhongcong Xu, Canbo Ye, Yinbai Hu, Hang Su, Zhengfa Liu, Guang Chen}
\IEEEauthorblockA{College of Automotive Engineering\\
Tongji University\\
Shanghai, China\\
}
}

\maketitle

\begin{abstract}
The neuromorphic camera is a brand new vision sensor that has emerged in recent years. In contrast to the conventional frame-based camera, the neuromorphic camera only transmits local pixel-level changes at the time of its occurrence and provides an asynchronous event stream with low latency. It has the advantages of extremely low signal delay, low transmission bandwidth requirements, rich information of edges, high dynamic range etc., which make it a promising sensor in the application of in-vehicle visual odometry system. This paper proposes a neuromorphic in-vehicle visual odometry system using feature tracking algorithm. To the best of our knowledge, this is the first in-vehicle visual odometry system that only uses a neuromorphic camera, and its performance test is carried out on actual driving datasets. In addition, an in-depth analysis of the results of the experiment is provided. The work of this paper verifies the feasibility of in-vehicle visual odometry system using neuromorphic cameras.
\end{abstract}


%
\IEEEpeerreviewmaketitle

\section{Introduction}
Neuromorphic vision sensors, also known as event cameras or dynamic vision sensors, capture pixel-level changes caused by movement in a scene (called "events"). Unlike conventional cameras, the output of a biological heuristic camera is not an intensity image but an event stream, which is continuous in time. An event can be represented by a tuple [t, x, y, p], where x and y represent the pixel coordinates of the events in the projected scene, t represents the timestamp and p indicates the polarity of the event (When the brightness is changed from dark to light, the polarity is positive; otherwise it is negative). Since the events are only generated when the brightness of a pixel changes, these pixels are mostly the edge of objects. This characteristic is advantageous for object recognition problem in computer vision because it can effectively reduce the storage and computational requirement. Moreover, event cameras can achieve an extremely high output frequency with a negligible latency within tens of microseconds. In addition, event cameras have a very high dynamic range of 130 dB, while conventional cameras can only achieve 60 dB\cite{Chen2018Neuromorphic}.

\section{Related Work}
Due to the characteristics of low latency, high dynamic range, sparse event stream and effective detection of the edges, the event cameras seem to be an ideal sensor for in-vehicle visual odometry system. Firstly, low latency ensures that the sensor equipped on the vehicle can obtain real-time information when the vehicle drives extremely fast. Secondly, high dynamic range improves the robustness against the extreme illumination change. Thirdly, the sparse event stream reduces the requirement of data transmission bandwidth.
\par In spite of the aforementioned advantages, it is still a tremendous challenge to apply the event camera on visual odometry system for vehicle since the output of the neuromorphic vision sensor is event stream, which is absolutely different with the conventional intensity images. Over the past few years, many researchers have attempted to solve this problem in more and more complex scenarios\cite{Rebecq2017EVO}. The complexity of these scenarios can be measured by the following three metrics: the type of motion, the type of scene and the dimensionality. 
\par The type of motion can be divided into rotational motion, planar motion (both are 3-DOF) and free motion in 3D space (6-DOF). Regarding the type of scene, it can be divided into artificial scenes and natural scenes. The artificial scene contains more texture information while the natural scene are much more complex because of larger illumination changes and more dynamic objects. In recent years, Some researchers introduce extra visual sensors such as conventional cameras or RGB-D cameras. In 2016, Rebecq et al.\cite{Rebecq2017EVO} has made a very detailed review about this area, so we optimize their review by adding some latest works on the basis of their review in Table 1.

\begin{table*}
\newcommand{\tabincell}[2]{\begin{tabular}{@{}#1@{}}#2\end{tabular}}
\caption{The relative works about the event camera based pose tracking and/or mapping.}
\centering
\begin{tabular}{cccccc}
	\toprule
	Reference& 2D/3D& Scene type& Event camera only& Depth& Remarks\\
	\midrule
	Cook et al. 2011 \cite{Cook2011Interacting}& 2D& natural& ${\surd}$&  X& rotational motion only\\
	\midrule
	Kim et al. 2014 \cite{BMVC_28_26}& 2D& natural& ${\surd}$& X& rotational motion only\\
	\midrule
	Weikersdorfer et al. 2013 \cite{Weikersdorfer2013Simultaneous}& 2D& artificial& ${\surd}$& X& a scene parallel to the plane of motion\\
	\midrule
	Censi et al. 2014 \cite{Censi2014Low}& 3D& artificial& X& X& requires depth fromextra RGB-D camera\\
	\midrule
	Weikersdorfer et al. 2014 \cite{Weikersdorfer2014Event}& 3D& natural& X& ${\surd}$& requires depth from extra RGB-D camera\\
	\midrule
	Mueggler et al. 2014 \cite{Mueggler2014Event}& 3D& artificial& ${\surd}$& X& requires a prior 3D map of lines\\
	\midrule
	Gallego et al. 2016 \cite{Gallego2016Event}& 3D& natural& X& X& requires a prior 3D map\\
	\midrule
	Kueng et al. 2016 \cite{Kueng2016Low}& 3D& natural& X& ${\surd}$& requires intensity images\\
	\midrule
	Kim et al. 2016 \cite{Kim2016Real}& 3D& natural& ${\surd}$& ${\surd}$& requires intensity images and GPU\\
	\midrule
	Rebecq et al. 2016 \cite{Rebecq2017EVO}& 3D& natural& ${\surd}$& ${\surd}$& \tabincell{c}{utilizes EMVS \cite{Rebecq2017EMVS} to reconstruct \\3D sparse point cloud}\\
	\midrule
	Gallego et al. 2017 \cite{Gallego2017Accurate}& 2D& natural& ${\surd}$& X& \tabincell{c}{estimates angular velocity with\\low latency and high precision}\\
	\midrule
	Zhu et al. 2017 \cite{Zhu2017Event}& 3D& natural& X& X& fusion with IMU based on Kalman filter\\
	\midrule
	Mueggler et al. 2018 \cite{Elias2017Continuous}& 3D& natural& X& ${\surd}$& fusion with IMU based on EVO \cite{Rebecq2017EVO}\\
	\midrule
	Rosinol et al. 2018 \cite{Vidal2018Ultimate}& 3D& natural& X& ${\surd}$& fusion with grayscale camera and IMU\\
	\bottomrule
\end{tabular}
\end{table*}

\par A relatively complete event based 2D SLAM system was proposed by Weikersdorfer et al. in \cite{Weikersdorfer2013Simultaneous}, which is limited to planar motion. In addition, this system requires a parallel plane with artificial black-and-white lines. A year later, Weikersdorfer et al.\cite{Weikersdorfer2014Event} extended their event based 2D SLAM system to 3D SLAM system with an extra RGB-D camera. The extra sensor provides depth information but it also slows down the SLAM system. In \cite{BMVC_28_26}, a filter based system was proposed for estimating the pose of the event camera and generating high resolution panorama, but this system is only able to estimate the rotational motion without transfer and depth. In \cite{Kueng2016Low} Kueng et al. proposed a visual odometry system for parallel tracking and mapping. The system estimates 6-DOF motion of camera in natural environment by tracking sparse features in event streams. This is the first event based visual odometry system that uses sparse feature points, but it still requires intensity images from conventional camera to detect feature points at first. Compared with the visual odometry system completely based on conventional camera, this system avoids a tremendous computational burden of feature tracking. In \cite{Kim2016Real} Kim et al. proposed a system based on three probabilistic filters to achieve pose tracking and reconstruction of depth and intensity images in natural scenes. This system is so computational expensive that it requires GPU, that makes it inappropriate for the platforms with limited computational power. EVO presented in \cite{Rebecq2017EVO} is the first 3D SLAM system only relies on event camera. It achieves parallel 6-DOF estimation and 3D map construction with a combination of the depth estimation algorithm from \cite{Rebecq2017EMVS} and a key-frame based visual odometry system. EVO matches the frames of short-term accumulation of events with the already built map to estimate the pose of camera. This matching procedure is computational efficient, so EVO can work in real time on CPU.
\par In recent years many researches has focused on the fusion of multiple sensors to achieve higher accuracy and robustness. In \cite{Zhu2017Event} and \cite{Elias2017Continuous}, the authors combined the event camera with IMU. In \cite{Vidal2018Ultimate}, Vidal et al. even combined three kind of sensors: the event camera, IMU and the grayscale camera, to provide a mature engineering application. 

\section{Event based visual odometry using feature tracking}
\begin{figure}[h]
	\centering
	\includegraphics[width=2.5in]{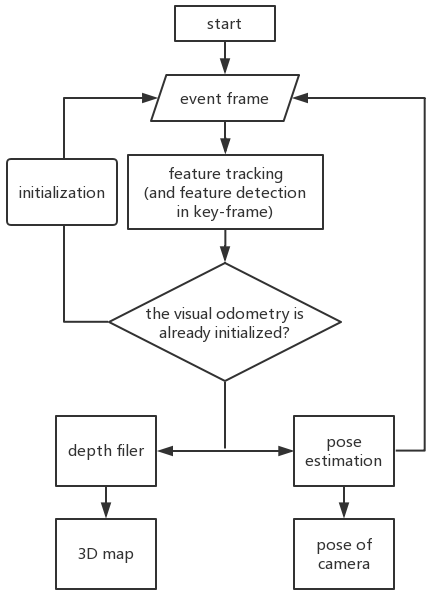}
	\caption{\label{framework} The framework of our event based visual odometry system}
\end{figure}

The main contribution of this paper is to propose a event based in-vehicle visual odometry system that utilizing the feature tracking algorithm proposed by Alex Zhu et al. in \cite{Zhu2017Event}. The framework of our visual odometry system is shown in \figurename{\ref{framework}}.

Our visual odometry system refers to the framework of some popular visual odometry system\cite{Forster2014SVO}\cite{Klein2008Parallel}\cite{Murartal2017ORB}\cite{Engel2014LSD}. It use a event camera to obtain asynchronous and low-latency event stream. With the tracked feature points and the already built 3D map, we can estimate the current pose of camera by solving PnP (Perspective-n-Point) problem\cite{Harltey2003Multiple}, which minimizes the reprojection error of feature points.
After that, the depth of feature points can be calculated by triangulation\cite{Maxwell2003Computer}. Due to the inevitable error of the measurement, we introduce depth filter\cite{Forster2014SVO}\cite{Vogiatzis2011Video} to model the depth estimation of the feature points with a probabilistic distribution. When the variance of the depth distribution is lower than the threshold after several updates, the corresponding point will be inserted into the 3D map.
\par It is noticeable that these two subsystems (pose estimation and depth estimation) rely on each other since the pose estimation needs the global 3D map while the update of depth filter needs the current camera pose. But there is no available map when our visual odometry system begins to work. For this reason, we estimate the camera pose by 2D-2D methods in initialization.
The 2D-2D pose estimation method that are most widely used is based on Homography\cite{Agarwal2005A}, but it is not appropriate for our target applicational scenario because most feature points in driving scenario are not on the same plane. So our visual odometry system utilizes Eight-Point-Algorithm\cite{Longuet1981A} for the essential matrix, which can then be used to reconstruct the rotation and transfer between the first two event frames.
\par In order to improve the computational efficiency and decrease the storage space, our visual odometry system only regularly stores an event frame as a key frame. When dealing with these key frames, our visual odometry system detects feature points and resets the depth filter while it only tracks the previous detected feature points and updates depth filter on ordinary frames. 

\subsection{Feature tracking}
\begin{figure*}[h]
	\centering
	\includegraphics[width=7in]{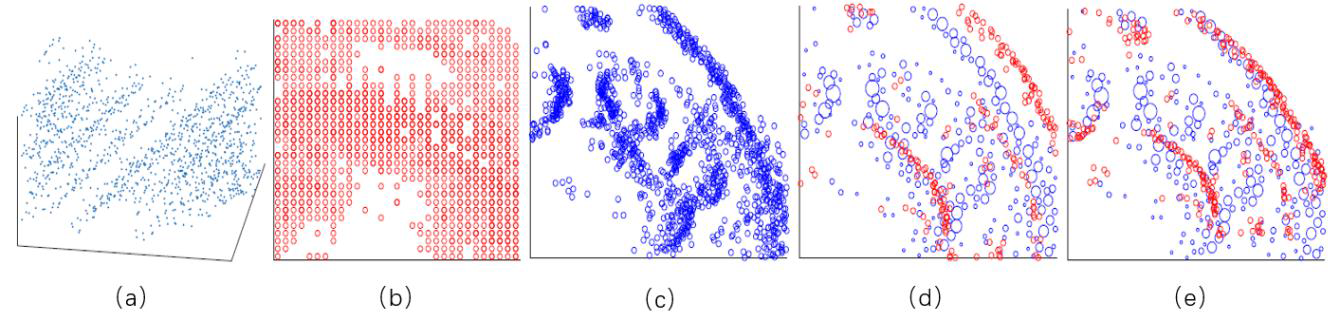}
	\caption{\label{alex_zhu} The procedure of feature tracking: (a) the ordinary event stream in spatiotemporal coordinate system; (b) events aggregated in a event frame without flow correction; (c) the corrected event frame; (d) two corrected event frames before affine warping; (e) the result of affine warping (from \cite{Zhu2017Event})}
\end{figure*}

The feature detection and feature tracking algorithm utilized in our system is proposed by Zhu et al. in \cite{Zhu2017Event}. \figurename{\ref{alex_zhu}} presents the process of this algorithm. First, the event points in a adjustable time interval are aggregated into a event frame. Secondly, this frame is corrected by Expectation Maximization optical flow estimation to propagate the events within a spatiotemporal window to the initial time t$_{0}$. The corrected image is similar to an edge map, then the feature points are detected by Harris corner detector on it. The final step is the feature alignment between the continuous corrected frames. \figurename{\ref{alex_zhu}}.(d) shows two continuous corrected event frames before affine warping and \figurename{\ref{alex_zhu}}.(e) shows the result of affine warping.

\subsection{Pose tracking}
Our visual odometry system utilizes bundle adjustment to estimate the camera pose through the minimization of reprojection error $e(\xi)$, which is defined by $$e(\xi)=u_{i}-\frac{1}{Z_{i}}K*exp(\xi^\land)P_{i} \eqno{(1)}$$
where $\xi$ is the camera pose ($\xi\in se(3), T\in SE(3)$), $u_{i}$ is the pixel coordinate of feature points in event frame, $Z_{i}$ is depth, K is camera intrinsics and $P_{i}$ is the coordinate of 3D point in the global map. The first order Taylor expansion of $e(\xi)$ is
$$e(\xi+\Delta\xi)\approx e(\xi)+J(\xi)\Delta\xi \eqno{(2)}$$
where J($\xi$) is the derivative of e($\xi$) and a 2×6 Jacobian matrix. Hence, the target $\Delta\xi$ is
$$\Delta\xi = \mathop{\arg\min}_{\Delta\xi} \frac{1}{2}\| e(\xi)+J(\xi)\Delta\xi \|^2_2\eqno{(3)}$$
The result of Eq.3 is when the derivation of $e(\xi)$ to $\Delta\xi$ is 0
$$J(\xi)^T J(\xi)\Delta\xi = -J(\xi)e(\xi)\eqno{(4)}$$
since there are $n$ feature points in the event frames, Eq.4 is transformed as follow
$$\sum_{i=1}^{n}w_i J(\xi)^T J(\xi)\Delta\xi = -\sum_{i=1}^{n}w_i J(\xi)e(\xi)\eqno{(5)}$$
where $w_i$ is the weight based on the precise of reprojection\cite{Kueng2016Low}. Since $J(\xi)$ and $e(\xi)$ of each feature points are both known, we can calculate $\Delta\xi$ from Eq.5 in iterations until $\Delta\xi$ is small enough.

\subsection{Mapping}
\begin{figure}[h]
	\centering
	\includegraphics[width=3.4in]{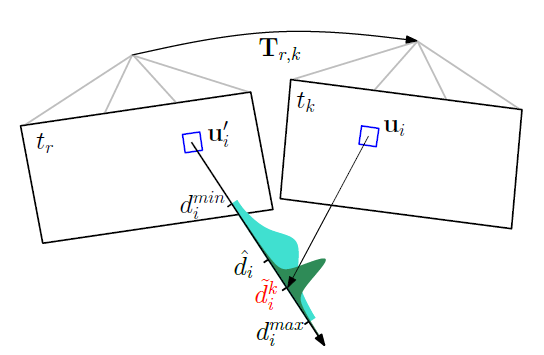}
	\caption{\label{depth_filter_update} Depth-filter update for a new measurement $\widetilde{d}^k_i$ at time $t_k$. The uncertainty of depth distribution becomes narrower.  (from \cite{Forster2014SVO})}
\end{figure}

In the last subsection we have estimated the pose of camera by bundle adjustment, then we can estimate depth of feature points by triangulation. Supposed that the homogeneous coordinates of a 3D point in two reference frames of camera are $x_1$ and $x_2$ respectively ($x_1=(u_1,v_1,1)^T$ and $x_2=(u_2,v_2,1)^T$). The geometry constraint between $x_1$ and $x_2$ is
$$Z_{2}x_{2} = Z_{1}Rx_{1}+t\eqno{(6)}$$
where $Z_{1}$ and $Z_{2}$ are depth, R and t are rotation and transfer between two event frames. After left multiplying $x_{2}^\land$ on both sides of Eq.6, the equation is transformed to
$$Z_{1}x_{2}^\land Rx_{1}+x_{2}^\land t = Z_{2}x_{2}^\land x_{2} = 0\eqno{(7)}$$
From Eq.7 we can calculate $Z_{1}$ and $Z_{2}$ for each feature point as a rough estimation. Because of the inevitable measurement error of depth estimation, we introduce depth filter to model the depth estimation of the feature points with a probability distribution. This depth filter was proposed in \cite{Vogiatzis2011Video}, which models the measurement $\widetilde{d}^k_i$ (k-th observation of the i-th feature) using a Gaussian+Uniform mixture distribution:
$$p(\tilde{d_i^k}|d_i,\rho_i)=\rho_i N(\tilde{d_i^k}|d_i,\tau^2_i)+(1-\rho_i)U(\tilde{d_i^k}|d^{min}_i,d^{max}_i)\eqno{(8)}$$
where $\rho_i$ is the inlier probability, which is close to 1 when the feature is well-tracked, $\tau_i^2$ is the variance caused by one-pixel disparity in the image plane and can be computed by geometric constraint. $[d^{min}_i, d^{max}_i]$ is the known range of depth estimation. The update for a new measurement $\widetilde{d}^k_i$ is shown in \figurename{\ref{depth_filter_update}}. And the detail of update of $d_i$ and $\rho_i$ is described in \cite{Vogiatzis2011Video}. When the uncertainty of the depth distribution is smaller than the threshold, the corresponding feature point will be inserted into the global map for further camera pose estimation.

\subsection{Parallel tracking and mapping}
The framework of our event based visual odometry system has been shown in \figurename{\ref{framework}}. The input is event stream consisted of series of event tuple [t, x, y, p]. In order to improve the computational performance, the feature points are only detected regular in the key frames, and the corresponding depth filter are reset. Regarding the ordinary event frames, our visual odometry system tracks the feature points to estimate the pose of camera and update the depth filter.
\par The pose estimation and the update of depth filter are handled by two main threads in our system. In pose estimation main thread, our visual odometry system utilizes bundle adjustment to estimate camera pose based on the tracked feature points in the current event frame and a already built map. In depth estimation main thread, our visual odometry system updates the depth filter with rotation and transfer between two consecutive frames. If the depth uncertainty of a feature point is lower than the threshold, this feature point will be inserted into the global map. The reason why our system works in two parallel threads is that pose tracking must be synchronous with the input of event camera while the real-time constraint of depth estimation main thread is not so strong relatively. Since the computational burden of depth estimation is relative greater, it's possible that a new event frame is received before the depth filters have been completely updated. In this case, our visual odometry system will store the pending event frames in a FIFO queue for the depth estimation.

\subsection{Bootstrapping}
Since two main threads of our visual odometry system (pose tracking and depth estimation) rely on each other but the visual odometry system has no available 3D global map for bundle adjustment when dealing with the first two event frames, our system utilizes Eight-point-algorithm\cite{H1981A} for essential matrix in bootstrapping. 
Supposed that the homogeneous coordinate of a 3D point in two reference frame of camera are $x_1$ and $x_2$ respectively ($x_1=(u_1,v_1,1)^T$ and $x_2=(u_2,v_2,1)^T$). The geometric constraint between $x_1$ and $x_2$ is 
$$Z_2x_2=Z_1Rx_1+t\eqno{(9)}$$
After left multiplying $x_{2}^Tt^\land$ on both sides of Eq.9, the equation is transformed to
$$x^T_2t^\land Rx_1=0\eqno{(10)}$$
$$\left(
\begin{array}{ccc}
u_2& v_2& 1
\end{array}
\right)
\left(
\begin{array}{ccc}
e_1& e_2& e_3\\
e_4& e_5& e_6\\
e_7& e_8& e_9\\
\end{array}
\right)
\left(
\begin{array}{c}
u_1\\v_1\\1
\end{array}
\right)
=0 \eqno{(11)}$$
where $t^\land R$ is a essential matrix and $e_i$ is the element of it (the depth can be neglected as a scalar multiplication). After converting the essential matrix to vector form, Eq.11 is transformed to
$$\begin{array}{ccccccccc}
(u_1u_2&  u_1v_2&  u_1&  v_1u_2& v_1v_2& v_1& u_2& v_2& 1)
\end{array}*e=0 \eqno{(12)}$$
where
$$e=
\begin{array}{cccc}
(e_1& e_2& ...& e_9)^T
\end{array}\eqno{(13)}
$$
There are 9 elements in the essential matrix, but since the essential matrix is defined only up to scale, if we can detect 8 pairs of feature points in the first two event frames, then we can obtain essential matrix to compute the rotation and transfer between the first two event frames.

\section{Experiment}

\subsection{Implementation details}
Our event based visual odometry system is implemented in C++ to take advantage of its object-oriented programming. There are several C++ classes in our visual odometry system: Config, Point, Feature, Frame, Map, PoseOptimizer, DepthFilter and FrameHandler. Config class is designated to read in the configuration file for some parameter. The Feature class represents the feature points in every event frame. When the depth uncertainty of a feature point is lower than the threshold, the corresponding Feature object will contain a pointer to a Point object, which represents the point in global map. The Frame class and the Map class are used to manage the Feature objects in the event frames and the Point objects in the global map respectively. In addition, the FrameHandler and DepthFilter are utilized for the calculation in pose estimation and depth estimation respectively. The FrameHandler class is the central of our visual odometry system, in which the PoseOptimizer class, the DepthFilter class and the Map class are initialized.

\subsection{Dataset}
Two datasets are used in this experiment. One is The Event-Camera Dataset and Simulator\cite{Mueggler2017The} provided by Robotics and Perception Group from the University of Zurich, the other one is Multi Vehicle Stereo Event Camera Dataset\cite{Zhu2018The} from Pennsylvania University.
\par
\begin{figure}[h]
	\centering
	\includegraphics[width=3.5in]{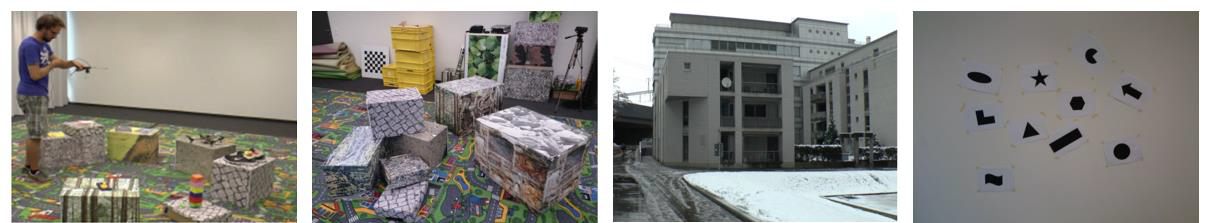}	
	\caption{\label{uzh}Some scenarios in Event-Camera Dataset and Simulator dataset}
\end{figure}

 Event-Camera Dataset and Simulator dataset was recorded by DAVIS camera (the version is DAVIS240C). Most of the data are recorded indoors by event camera, conventional camera and IMU, and this dataset also provide the ground truth of camera pose. As for the outdoor data, a precise pose estimation is provided by IMU. This dataset is recorded on hand-held device or drone instead of vehicle, that is different from our target application scenario, but this dataset provides two file formats of recording: txt and rosbag, and the txt-format data provides dramatic convenience for the debug phase of our system. \figurename{\ref{uzh}}shows some scenarios in this dataset.
\par The Multi Vehicle Stereo Event Camera Dataset was recorded by different sensors (event camera, conventional camera, 16-beam LiDAR, GPS, IMU) equipped on different platforms (car, motor cycle, drone). The data recorded by event camera equipped on vehicle are mostly from the urban roads at both day and night, and the pose of camera is calculated by localization algorithm based on LiDAR, IMU and GPS. This dataset includes several driving scenarios such as start-up acceleration, constant speed driving, deceleration at crossing and parking with great changes of illumination and the noise introduced by dynamic objects. 
\par Additionally, our experiment is carried out on ROS Melodic. Open source libraries such as Eigen, Sophus, OpenCV, g2o, Boost are also utilized in our visual odometry system.

\subsection{Feature detection and tracking}
The experiment of feature detection and tracking is carried out in scenarios of both daytime and night in Multi Vehicle Stereo Event Camera Dataset. 
\par
\begin{figure}[h]
	\centering
	\includegraphics[width=3.4in]{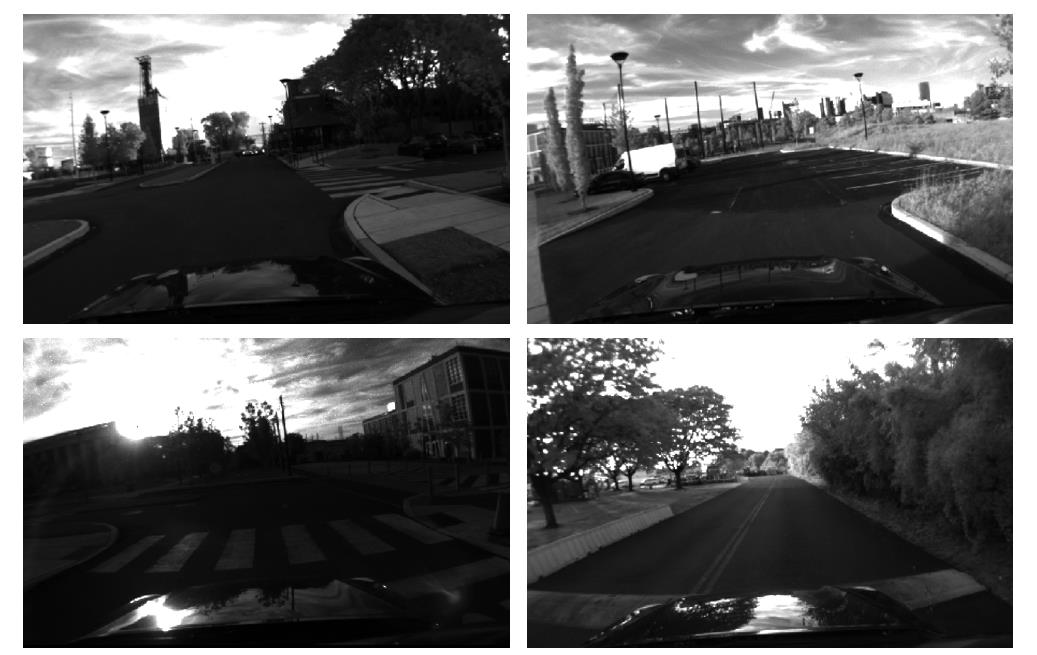}	
	\caption{\label{daytime1} The grayscale images of daytime in Multi Vehicle Stereo Event Camera Dataset}
\end{figure}

\begin{figure}[h]
	\centering
	\includegraphics[width=3.4in]{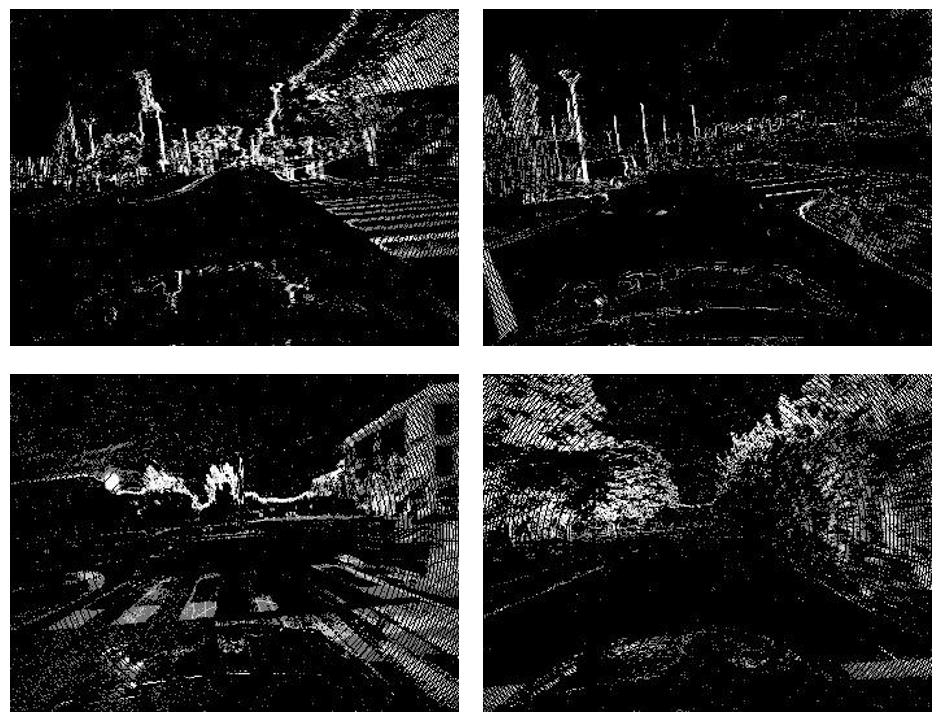}	
	\caption{\label{daytime2}The event frames of daytime in Multi Vehicle Stereo Event Camera Dataset}
\end{figure}

\begin{figure}[h]
	\centering
	\includegraphics[width=3.4in]{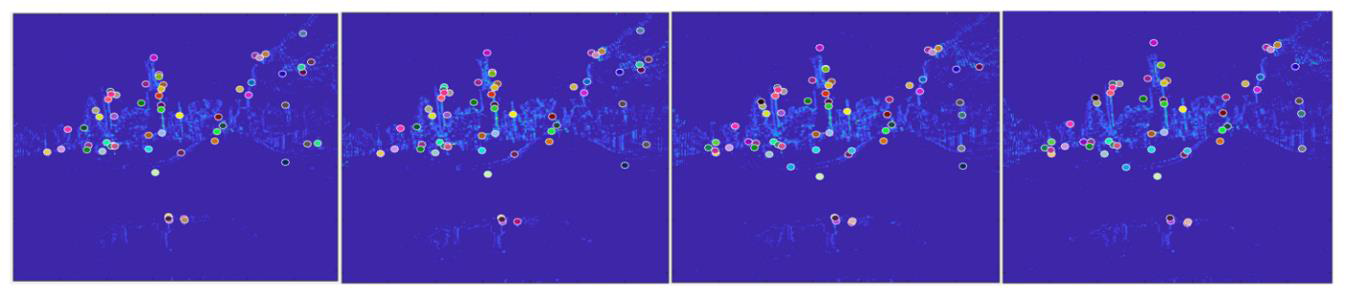}	
	\caption{\label{daytime3}The feature tracking in four continuous daytime event frames (the same feature points are represented by the same color)}
\end{figure}

 Some grayscale images and event frames of daytime scenario are shown in \figurename{\ref{daytime1}} and \figurename{\ref{daytime2}}. Thanks to high dynamic range of event cameras, the event frames show the robustness against the dramatic illumination change and partial overexposure. The \figurename{\ref{daytime3}} shows the feature tracking in four continuous event frames.

\begin{figure}[h]
	\centering
	\includegraphics[width=3.4in]{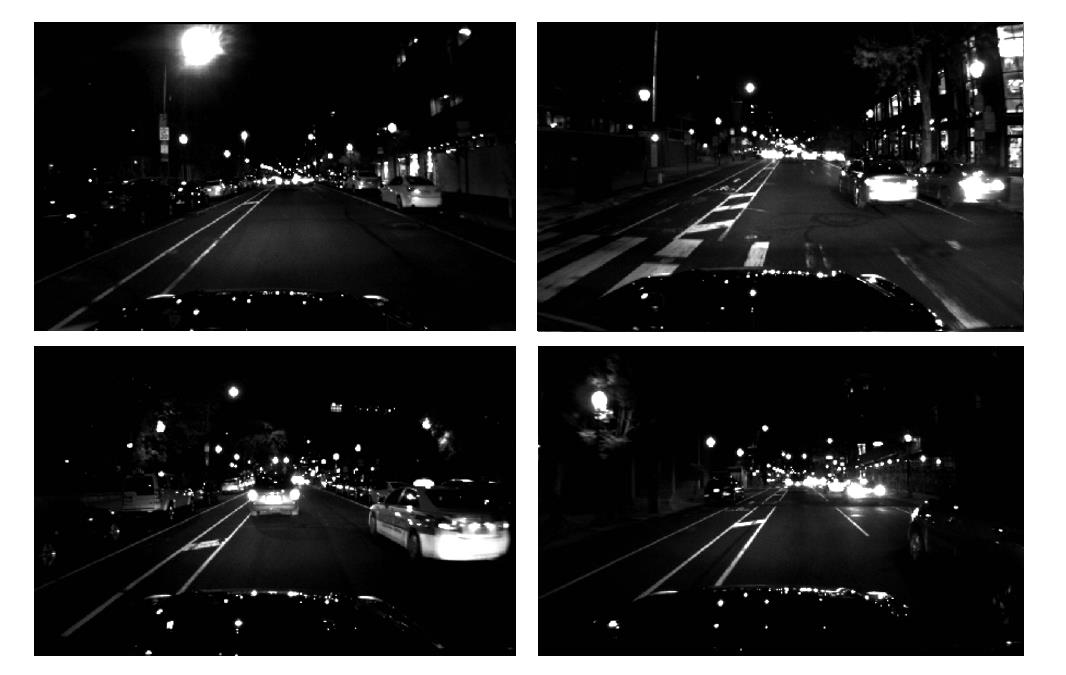}	
	\caption{\label{night1} The grayscale images of night in Multi Vehicle Stereo Event Camera Dataset}
\end{figure}

\begin{figure}[h]
	\centering
	\includegraphics[width=3.4in]{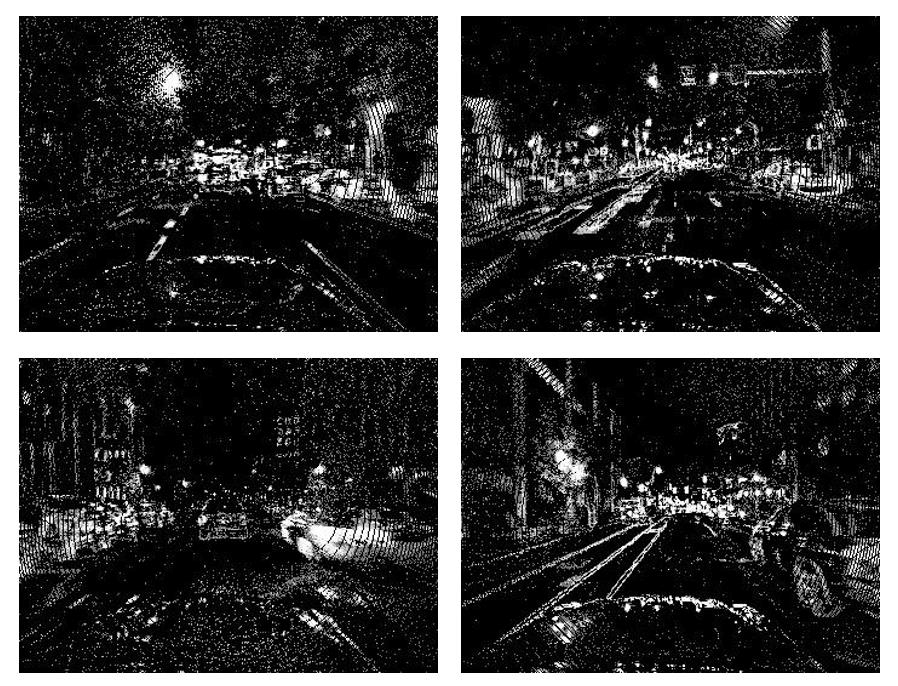}	
	\caption{\label{night2}The event frames of night in Multi Vehicle Stereo Event Camera Dataset}
\end{figure}

\begin{figure}[h]
	\centering
	\includegraphics[width=3.4in]{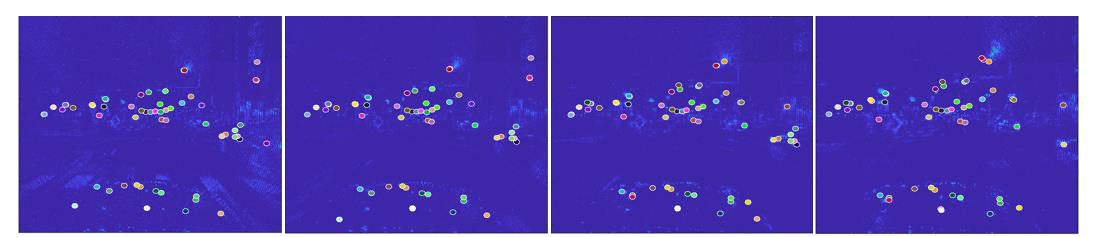}	
	\caption{\label{night3}The feature tracking in four continuous night event frames (the same feature points are represented by the same color)}
\end{figure}

Some grayscale images and event frames of night scenario are shown in \figurename{\ref{night1}} and \figurename{\ref{night2}}. Due to the poor illumination at night, it is quite hard to distinguish the vehicles and buildings in the dark. In addition, artificial lights cause partial overexposure. These problems are alleviated in the output of event camera, but the data recorded from the event camera contains relatively much more noise, which is a main drawback of the current event camera. \figurename{\ref{night3}} shows the feature tracking in four continuous event frames.

\par
\begin{figure}[h]
	\centering
	\includegraphics[width=3.5in]{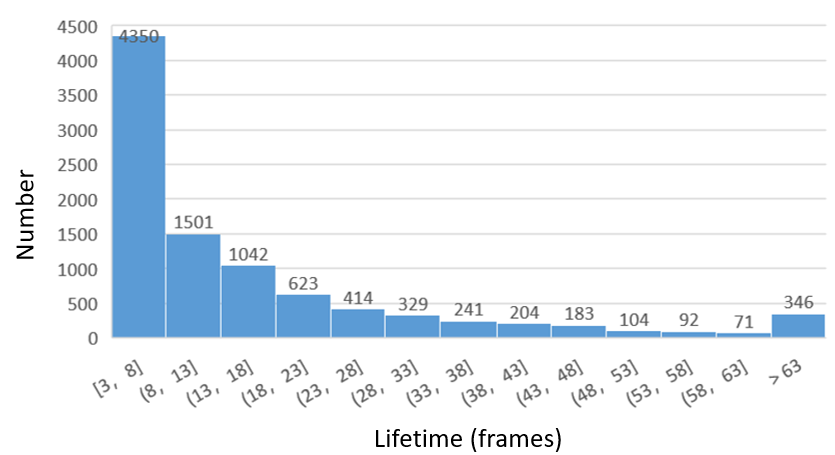}	
	\caption{\label{distribution_daytime}The distribution of lifetime in daytime scenario}
\end{figure}

\begin{figure}[h]
	\centering
	\includegraphics[width=3.5in]{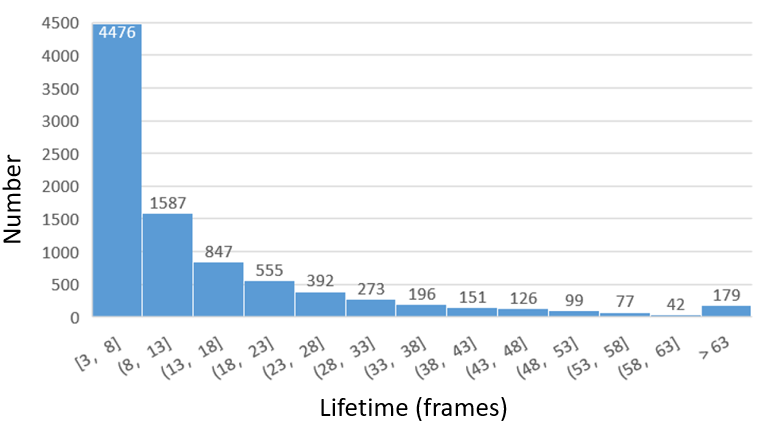}	
	\caption{\label{distribution_night}The distribution of lifetime in night scenario}
\end{figure}

 The stability of feature tracking is an important criterion in visual odometry system, because the variance of depth distribution can not converge unless feature points can be tracked in enough continuous event frames. And the feature detection is carried out only in key event frames, so a feature point can not be tracked once it is lost in a ordinary event frame. We adjust the stability of our system using 'lifetime'\cite{Mueggler2015Lifetime}, which is the span of event frames from a feature point is detected until it is lost. The statistics result is shown in Table 2 (the feature points whose lifetime is less than 3 event frames will be ignored) and the distribution of lifetime in daytime and night scenarios are shown in \figurename{\ref{distribution_daytime}} and \figurename{\ref{distribution_night}} respectively.
\begin{table}
	\caption{The average lifetime of feature points in daytime and night scenarios}
	\centering
	\begin{tabular}{ccc}
		\toprule
		Item& Daytime& Night\\
		\midrule
		Average lifetime& 16.488 frames& 14.321 frames\\
		Standard variance& 18.505 frames& 15.687 frames\\
		\bottomrule
	\end{tabular}
\end{table}

The average lifetime of feature points in daytime and night scenarios are 16.488 frames and 14.321 frames respectively.  Considering that the variance of depth filter can converge after 5-7 iterations, the stability of feature tracking meets the requirement generally. It can also be seen that the difference between the tracking performance in daytime and night scenarios is much smaller than that of conventional visual odometry system, which proves that the event camera can detect the edges efficiently even under poor illumination condition. But the real-time performance of this feature tracking algorithm is quite pool. Using the source code provided by Alex Zhu et al., our visual odometry system is far from meeting the requirement of tracking about 100 feature points synchronously, so until now our system can only run in offline mode. In \cite{Zhu2017Event}, Alex Zhu et al. claimed that they are improving the real-time performance of the algorithm to track more feature points synchronously, we will continuously follow their research project. 

\subsection{Event based visual odometry system}
\begin{figure}[h]
	\centering
	\includegraphics[width=3.5in]{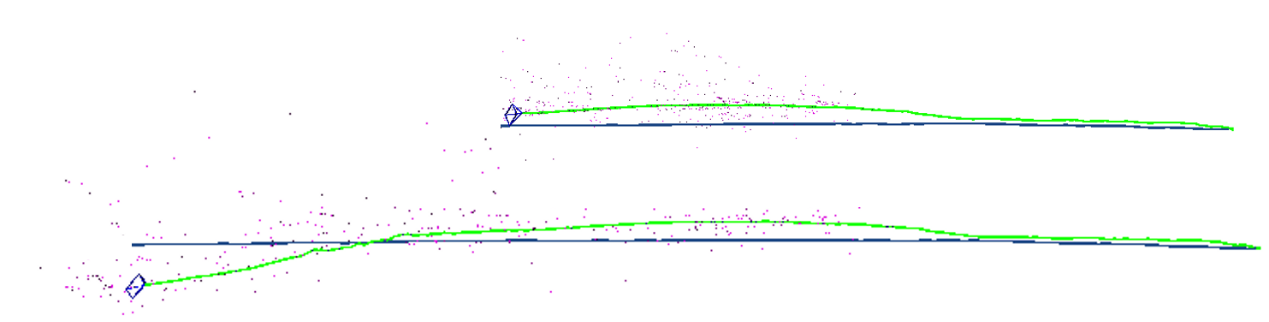}	
	\caption{\label{odometry_result_1}The localization estimation from our event based visual odometry system and the ground truth (the blue line is the ground truth provided by LOAM and the green line is the trajectory estimated by our visual odometry system)}
\end{figure}

Since the real-time performance of the feature tracking has not been implemented, we first record the result of feature tracking (the identifier of feature points, the coordinate in event frames) in a txt-format file and then utilize this file in our odometry experiment. In Multi Vehicle Stereo Event Camera dataset, the localization is provided by GPS and other algorithms like LOAM and Cartographer, and we finally choose the localization provided by LOAM, which is based on IMU, as the ground truth. \figurename{\ref{odometry_result_1}} shows the visualization of the estimated trajectory in our experiment.
\par
\begin{figure}[h]
	\centering
	\includegraphics[width=3.5in]{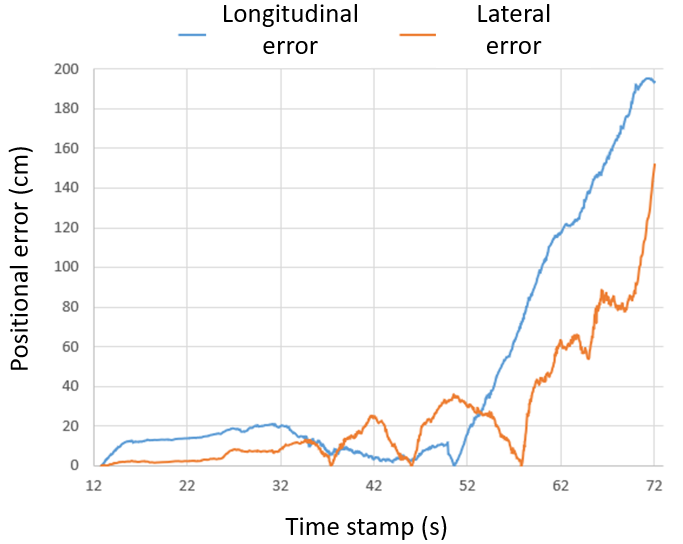}	
	\caption{\label{odometry_result_2}The localization error of our event basd visual odometry system in the interval (from 12s to 72s)}
\end{figure}

 The timestamp of the event frames used in this experiment is from 12s to 72s and the driving distance in this interval is about 439 meters. The localization error in this interval is shown in \figurename{\ref{odometry_result_2}} and \tablename{\ref{table_3}}.
\begin{table}
	\caption{\label{table_3}The statistics result of positional error}
	\centering
	\begin{tabular}{cccc}
		\toprule
		Item& Longitudinal& Lateral& Planar\\
		\midrule
		Average error& 0.488 m& 0.275 m& 0.581 m\\
		Relative error& 0.4399\%& 0.3464\%& 0.5578\%\\
		\bottomrule
	\end{tabular}
\end{table}
The average error of planar localization is 0.581 m, which is larger than the most visual odometry system based on the conventional cameras. Additionally, the relative position error of our odometry system is 0.5578\% while the relative position error of EVO\cite{Rebecq2017EVO} system is only 0.2\%. In \figurename{\ref{odometry_result_2}} we can see that the longitudinal and lateral positional error are kept within 0.2 meter before 52s but the position error expands dramatically after 52s since the vehicle passed two bumps.

\section{Conclusion}
In this paper we design a event based visual odometry system that utilizes feature tracking to achieve parallel pose estimation and mapping. We test our system on dataset about urban road scenario and the result shows that our system has a good performance on feature tracking and robustness against extreme illumination changes and dark scenarios. But the real-time performance of our visual odometry system is poor because of the huge computational burden of feature tracking. In addition, the localization error of our system increase sharply when the vehicle passes a bump. 
But due to the low latency, the effective detection of edges and other aforementioned advantages, event camera is still a promising sensor for autonomous driving. For this reason, we will keep improving the real-time performance and the robustness against bumpy road of our event based visual odometry system.

\section*{Acknowledgment}
The research leading to these results has partially received funding from the Shanghai Automotive Industry Sci-Tech Development Program (Grant. No. 1838), from the Shanghai AI Innovative Development Project 2018, and from the Young Scientists Fund of the National Natural Science Foundation of China (Grant No. 6190021037).



%
\bibliographystyle{IEEEtran}      

\end{document}